\documentclass{article}

\usepackage[main, preprint]{mystyle}

\usepackage[utf8]{inputenc} 
\usepackage[T1]{fontenc}    
\usepackage{url}            
\usepackage{booktabs}       
\usepackage{amsfonts}       
\usepackage{nicefrac}       
\usepackage{microtype}      
\usepackage{xcolor}         
\usepackage{subcaption}
\usepackage{setspace}
\usepackage{graphicx}
\usepackage{multirow}
\usepackage{amsmath}
\usepackage{adjustbox}
\usepackage{caption}
\usepackage{enumitem}
\usepackage{makecell}
\usepackage{colortbl}
\usepackage{wrapfig}
\usepackage{pifont}
\usepackage[most]{tcolorbox}

\usepackage{algorithm}          
\usepackage{listings}           
\usepackage{marvosym}
\usepackage{bxcoloremoji}

\setlist[itemize]{leftmargin=*}

\title{\textit{SwimBird}: Eliciting \underline{Swi}tchable Reasoning \underline{M}ode in\\Hy\underline{brid} Autoregressive MLLMs}

\author{
	\textbf{Jintao Tong}$^{1,2}$\quad \textbf{Shilin Yan}$^{2\dagger}$\textsuperscript{\ddag} \quad \textbf{Hongwei Xue}$^2$\quad \textbf{Xiaojun Tang}$^{2}$\quad\textbf{Kunyu Shi}$^{2}$ \and  \textbf{Guannan Zhang}$^{2}$\quad \textbf{Ruixuan Li}$^{1}$\textsuperscript{\ddag} \quad \textbf{Yixiong Zou}$^{1}$\textsuperscript{\ddag}
	\and \\	
	$^1$Huazhong University of Science and Technology~~
	$^2$Accio Team, Alibaba Group \and
	$^{\dagger}$ Project Leader \quad \textsuperscript{\ddag} Corresponding Author
}

\headerlogos{
	\includegraphics[height=0.5cm]{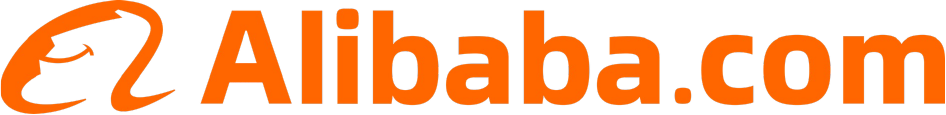}\hspace{0.3cm}
	\includegraphics[height=0.5cm]{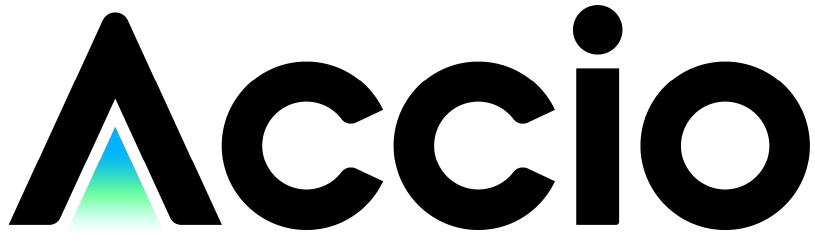}  
}

\newcommand{\github}{\raisebox{-1.5pt}{\includegraphics[height=1.05em]{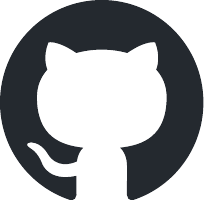}}}

\begin{document}
	
\maketitle
\vspace{-0.5cm}
\begin{abstract}
Multimodal Large Language Models (MLLMs) have made remarkable progress in multimodal perception and reasoning by bridging vision and language. However, most existing MLLMs perform reasoning primarily with textual CoT, which limits their effectiveness on vision-intensive tasks. Recent approaches inject a fixed number of continuous hidden states as “visual thoughts” into the reasoning process and improve visual performance, but often at the cost of degraded text-based logical reasoning. We argue that the core limitation lies in a rigid, pre-defined reasoning pattern that cannot adaptively choose the most suitable thinking modality for different user queries. We introduce SwimBird, a reasoning-switchable MLLM that dynamically switches among three reasoning modes conditioned on the input: (1) text-only reasoning, (2) vision-only reasoning (continuous hidden states as visual thoughts), and (3) interleaved vision–text reasoning. To enable this capability, we adopt a hybrid autoregressive formulation that unifies next-token prediction for textual thoughts with next-embedding prediction for visual thoughts, and design a systematic reasoning-mode curation strategy to construct SwimBird-SFT-92K, a diverse supervised fine-tuning dataset covering all three reasoning patterns. By enabling flexible, query-adaptive mode selection, SwimBird preserves strong textual logic while substantially improving performance on vision-dense tasks. Experiments across diverse benchmarks covering textual reasoning and challenging visual understanding demonstrate that SwimBird achieves state-of-the-art results and robust gains over prior fixed-pattern multimodal reasoning methods. \\

\coloremojicode{1F310}  \textbf{Project Page:} \href{https://accio-lab.github.io/SwimBird}{https://accio-lab.github.io/SwimBird} \vspace{0.1cm}

\github{}  \textbf{Github Repo:} \href{https://github.com/Accio-Lab/SwimBird}{https://github.com/Accio-Lab/SwimBird} \vspace{0.1cm}

\coloremojicode{1F917}  \textbf{HuggingFace:} \href{https://huggingface.co/datasets/Accio-Lab/SwimBird-SFT-92K}{https://huggingface.co/datasets/Accio-Lab/SwimBird-SFT-92K}

\end{abstract}

\section{Introduction}
Building on the success of Chain-of-Thought (CoT)~\cite{wei2022chain,kojima2022large} reasoning in LLMs, recent multimodal research has adopted step-by-step reasoning to decompose complex vision-and-language problems into intermediate steps that are easier to solve. With textual CoT, Multimodal Large Language Models (MLLMs)~\cite{zhang2023multimodal,huang2025vision,liu2025visual,tong2025emosync} have significantly improved on tasks requiring symbolic manipulation, numerical calculation, and logical analysis.

However, this success does not fully transfer to vision-dense tasks where the bottleneck lies in dense perception and spatial reasoning rather than logical structure~\cite{fu2024blink}. Typical examples include maze solving, fine-grained visual search, and other problems where accurate intermediate visual states are essential. On such tasks, purely textual CoT~\cite{shen2025vlm} can be an ill-posed interface: the model is forced to describe intermediate visual evidence in language even when language is not a faithful carrier, causing brittle reasoning and error accumulation~\cite{yu2025perception}. To address this, recent works introduce latent visual reasoning~\cite{li2025latent,tong2025sketch} that supervises models to generate semantically grounded continuous hidden states as visual thoughts, enabling intermediate visual representations to be maintained and updated across steps,  which substantially strengthens performance on vision-dense benchmarks.

\begin{figure*}[!t]
	\centering
	\includegraphics[width=\linewidth]{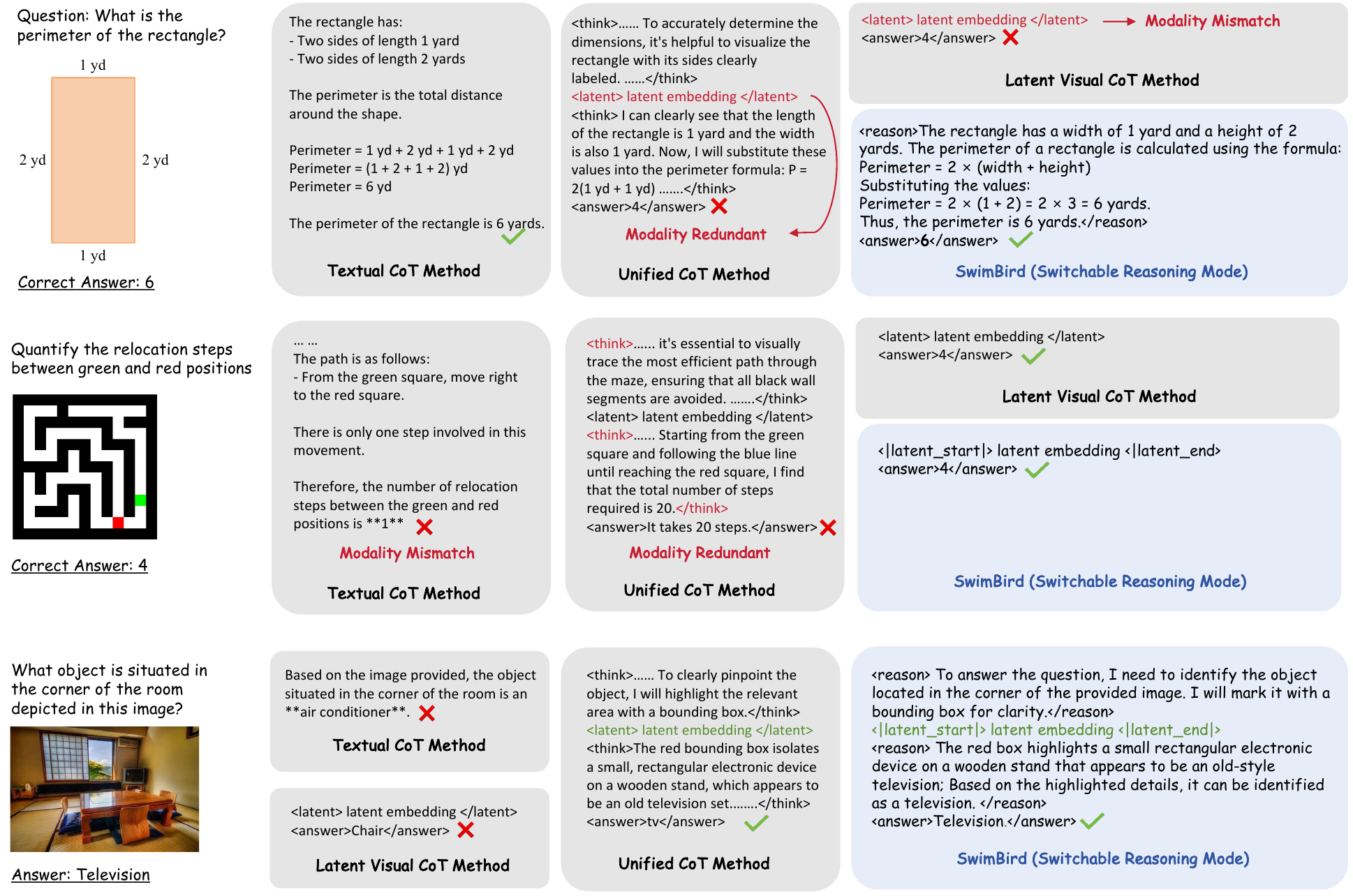} 
	\caption{\textbf{SwimBird enables query-adaptive multimodal reasoning by dynamically switching among text-only, vision-only, and interleaved vision--text modes}. As illustrated, it avoids redundant latent steps on text-centric queries (Case~1), relies on latent visual thoughts for vision-dense spatial problems (Case~2), and interleaves visual grounding with textual deduction when both are needed (Case~3), mitigating modality mismatch and improving robustness.}
	\label{Fig.intro}
\end{figure*} 

Despite these advances, existing multimodal CoT designs largely rely on a rigid, pre-defined reasoning pattern. Concretely, prior methods~\cite{wang2025vl,zhang2025latent,yang2025machine} typically fall into three fixed paradigms: text-only CoT, vision-only CoT, or interleaved vision–text CoT. As shown in Fig. 1, such fixed patterns create a mismatch between the reasoning modality and the actual needs of the question: forcing visual thoughts for text-centric queries can interfere with discrete symbolic reasoning, while restricting strongly visual queries to text-only reasoning removes an appropriate latent workspace. Even interleaved reasoning remains a fixed schedule that may generate redundant modality steps~\cite{tong2025flowcut}.

We argue that the core limitation is the assumption that a single, static reasoning template can generalize across heterogeneous multimodal queries. Different questions demand different internal computation formats. Some require only discrete symbolic steps, some require only latent visual transitions, and some require tight alternation between visual grounding and textual deduction. \textbf{\textit{A more capable MLLM should therefore be able to choose when to think in language, when to think in vision, conditioned on the input and the evolving reasoning state.}}

Motivated by this, we propose \textbf{SwimBird}, a reasoning-switchable MLLM for query-adaptive multimodal reasoning. SwimBird is built on two key ideas derived from the limitations above.
First, we adopt a hybrid autoregressive formulation that supports both (i) standard next-token prediction for textual thoughts and (ii) next-embedding prediction for continuous visual thoughts. This unified generation interface provides the foundation for switchable reasoning.
Second, we attribute the rigidity of prior patterns partly to training data bias.
We therefore design a systematic curation strategy that filters and categorizes multimodal CoT samples into reasoning modes based on their visual dependency and reasoning characteristics. Through this strategy, we construct \textbf{SwimBird-SFT-92K}, a diverse supervised fine-tuning dataset covering text-only, vision-only, and interleaved vision--text patterns. With these designs, SwimBird can dynamically switch among three reasoning modes.

Importantly, SwimBird also removes the fixed-budget constraint in visual reasoning. Instead of generating a constant-length sequence of visual thought tokens, it dynamically determines the number of visual thought tokens during vision-only or interleaved reasoning, allocating more latent computation to vision-dense queries while avoiding redundant visual thoughts for text-centric problems. As a result, a single model can robustly handle diverse query types, whereas fixed-pattern baselines typically excel only on a subset and may underperform when the required thinking modality or visual-thought budget deviates from their pre-defined design.

Our contributions are summarized as follows:
\vspace{-0.3cm}
\begin{itemize}[leftmargin=*]
	\setlength{\itemsep}{1pt}
	\setlength{\parskip}{1pt}
	\item We identify two key bottlenecks of prior multimodal CoT frameworks, namely fixed reasoning-mode templates and fixed visual-thought lengths, and show how they lead to a modality mismatch that harms either vision-dense performance or text-based logical reasoning.
	
	\item We introduce SwimBird, a hybrid autoregressive MLLM that can dynamically switch among text-only, vision-only, and interleaved reasoning modes, combining next-token prediction for textual thoughts with next-embedding prediction for visual thoughts.
	\item We further introduce adaptive visual-thought allocation, enabling SwimBird to dynamically determine the number of continuous visual-thought tokens based on query complexity.
	\item We design a systematic reasoning-mode curation strategy for multimodal CoT samples and construct SwimBird-SFT-92K, a dataset covering three reasoning patterns that enables query-adaptive mode selection.
	\item Extensive experiments across diverse benchmarks demonstrate that SwimBird achieves state-of-the-art performance on both text-centric reasoning and challenging vision-dense tasks, outperforming prior fixed-pattern multimodal reasoning methods.
\end{itemize}

\section{Related Works}

\subsection{Textual CoT in MLLMs}
The integration of vision and language has evolved from discriminative tasks toward generative reasoning frameworks. Early MLLMs focus primarily on visual question answering through direct answer generation~\cite{li2023blip, liu2023visual,wang2024qwen2,liu2024llavanext,yan2025crosslmm}. With the success of step-by-step reasoning in LLMs, recent MLLMs incorporate explicit reasoning chains to handle complex multimodal problems~\cite{bai2025qwen2,wang2025internvl3,xu2025llava}. These models generate intermediate textual explanations before producing final answers, demonstrating improved performance on mathematical word problems, scientific diagram understanding, and multi-hop visual reasoning~\cite{yue2024mmmu,wang2024charxiv,qiao2025we}. Despite their effectiveness on logic-heavy benchmarks, these text-based reasoning approaches struggle when the core challenge lies in visual perception rather than logical decomposition~\cite{shen2025zoomeye}. Tasks requiring spatial transformation tracking, visual state prediction, or fine-grained visual comparison expose the fundamental limitation that the model is forced to describe intermediate visual evidence in language, even when language is not a faithful or efficient carrier for the required information, leading to brittle reasoning and error accumulation.

\subsection{Latent Visual Reasoning}
Recognizing the constraints of language-only reasoning, researchers have explored alternative computational substrates for visual thinking~\cite{qin2025chain,wang2025monet}. Recent methods propose latent visual reasoning by training models to produce continuous embeddings supervised by visual reconstruction objectives. For instance, Mirage~\cite{yang2025machine} employs hidden states trained to approximate annotated helper images, while LVR~\cite{li2025latent} focuses on reconstructing cropped image regions. SkiLa~\cite{tong2025sketch} proposes unified reasoning that alternates between generating latent visual tokens and discrete textual tokens. However, existing latent reasoning methods uniformly apply the same reasoning structure across all inputs: models trained with visual thoughts always generate them, even for purely textual queries. Furthermore, these methods use fixed-length latent tokens regardless of whether a problem requires minimal or extensive visual deliberation. SwimBird addresses both limitations through dynamic mode selection and adaptive visual token budgets, enabling truly query-adaptive multimodal reasoning.

\begin{figure*}[!t]
	\centering
	\includegraphics[width=1\linewidth]{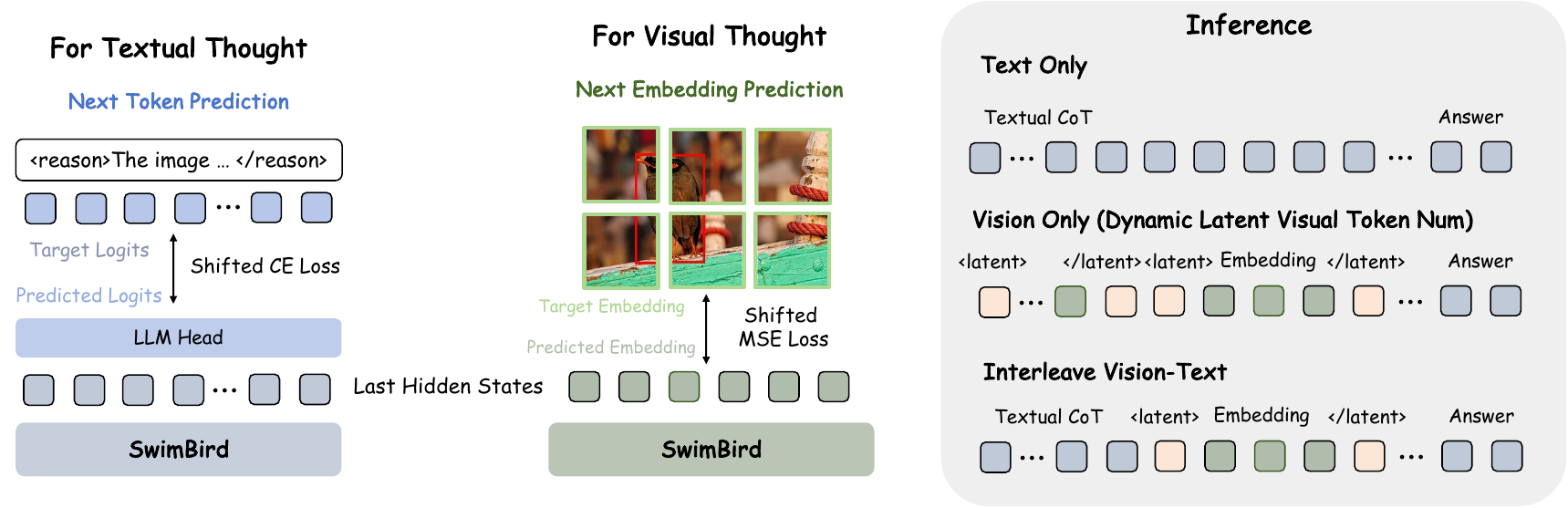}
	\caption{\textbf{SwimBird adopts a hybrid autoregressive formulation that performs next-token prediction for textual thoughts and switches to next-embedding prediction for visual thoughts}. During inference, SwimBird performs query-adaptive multimodal reasoning by dynamically selecting among three modes conditioned on the input: text-only, vision-only, and interleaved vision-text reasoning.}
	\label{Fig.method} 
\end{figure*} 

\section{Method}
SwimBird adopts a hybrid autoregressive formulation that supports both discrete textual tokens and continuous latent visual tokens. As shown in Fig.~\ref{Fig.method} (left), it performs standard next-token prediction for textual thoughts, optimized with a shifted cross-entropy loss, and performs next-embedding prediction for visual thoughts, optimized with a MSE loss to reconstruct the embeddings of intermediate thinking images.
During inference (Fig.~\ref{Fig.method} right), SwimBird performs query-adaptive reasoning by generating either (i) text-only traces, (ii) vision-only traces with a variable-length latent span, or (iii) interleaved vision--text traces, conditioned on the input.

\subsection{Hybrid Autoregressive Modeling}

\textbf{Textual thought as next-token prediction.}
For textual reasoning spans, SwimBird behaves like a standard language model. Given a token sequence $\{w_1,\dots,w_T\}$, the model outputs logits parameterizing
\begin{equation}
	p_\theta(w_t \mid w_{<t}, \mathbf{x}),
\end{equation}
where $\mathbf{x}$ denotes the observed image (and prior context). We train these spans with the standard cross-entropy loss:
\begin{equation}
	\mathcal{L}_{\text{text}} = - \sum_{t=1}^{T} \log p_\theta(w_t \mid w_{<t}, \mathbf{x}).
\end{equation}
This objective preserves the discrete symbolic manipulation and logical consistency of the language backbone, which is essential for text-centric reasoning tasks.

\textbf{Visual thought as next-embedding prediction.}
For vision-only reasoning or visual segments inside interleaved reasoning, SwimBird generates a sequence of continuous latent tokens (visual thoughts) $\{\mathbf{z}_1,\dots,\mathbf{z}_K\}$, each represented as a hidden-state embedding rather than a discrete word. Concretely, we treat each visual-thought step as predicting the next embedding in an autoregressive manner:
\begin{equation}
	\hat{\mathbf{z}}_k = f_\theta(\mathbf{z}_{<k}, w_{\le T}, \mathbf{x}),
\end{equation}
and supervise it with a shifted mean squared error (MSE) loss against target embeddings $\mathbf{z}_k$:
\begin{equation}
	\mathcal{L}_{\text{vis}} = \sum_{k=1}^{K} \left\lVert \hat{\mathbf{z}}_k - \mathbf{z}_k \right\rVert_2^2 .
\end{equation}
Here, the target embeddings are computed by encoding the intermediate thinking images with the same vision encoder (and projection) used by SwimBird, thus grounding latent visual thoughts in semantically meaningful visual states.

\textbf{Unified training objective.}
A training instance may contain pure textual CoT, pure visual CoT, or interleaved segments. We optimize a unified objective that sums modality-specific losses over the activated segments:
\begin{equation}
	\mathcal{L} = \lambda_{\text{text}} \mathcal{L}_{\text{text}} + \lambda_{\text{vis}} \mathcal{L}_{\text{vis}},
\end{equation}
where $\lambda_{\text{text}}$ and $\lambda_{\text{vis}}$ are balancing coefficients. In practice, each sample only contributes to the losses of the modes it contains, enabling the model to learn all three reasoning patterns without forcing unnecessary supervision.

\textbf{Mode switching with special delimiters}
To enable controllable and learnable switching among reasoning modes, we introduce explicit delimiters in the target sequences. Specifically, we mark visual-thought spans using special tokens such as \texttt{<|latent\_start|>} and \texttt{<|latent\_end|>}. During training, these delimiters define where the model should produce continuous latent embeddings instead of textual tokens. During inference, SwimBird generates these delimiters autoregressively, which makes mode selection \emph{query-adaptive}: the model can decide whether to enter a latent visual-thinking phase, remain in text-only reasoning, or alternate between the two (Fig.~\ref{Fig.method} right).

\begin{table*}[!t]
	\centering
	\setstretch{1.1}
	\resizebox{\linewidth}{!}{
		\begin{tabular}{lrrrrc}
			\toprule
			
			Data Source & All Mode & Text Only &Vision Only &Interleave & Problem Domain\\
			\midrule
			
			Zebra-CoT &26.3K  &0 &5.9K &20.4K  &Visual Search, Jigsaw, Maze, Geometry, Chess...\\
			ThinkMorph & 7.1K &0 &1.2K &5.9K  &Visual Search, Spatial Navigation, Jigsaw, Chart\\
			MathCanvas & 8.9K &0  &1.7K&7.2K &Geometry, Algebra, Calculus, Statistics\\
			OpenMMReasoner &50K &50K&0&0  &General VQA, Math VQA, Text QA\\
			\midrule
			Total &92.3K &50K &8.8K &33.5K &\\
			\bottomrule
		\end{tabular}
	}
	\caption{Detailed statistics of SwimBird-SFT-92K.}
	\label{tab:dataset} 
\end{table*}

\subsection{Dynamic Latent Token Budget}

\begin{wrapfigure}{r}{0.49\textwidth} 
	\vspace{-0.6cm}
	\centering
	\includegraphics[width=\linewidth]{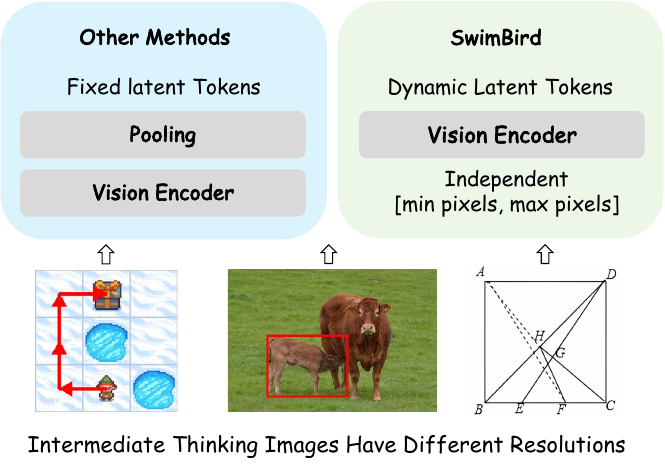} \vspace{-0.3cm}
	\caption{\textbf{Resolution-aware, dynamic latent tokens budget.}}
	\label{Fig.dynamic}
	\vspace{-0.2cm}
\end{wrapfigure}

Prior latent visual reasoning methods typically adopt a fixed number of latent tokens (or a fixed pooling strategy) for all inputs. This design has two drawbacks: (1) it can lead to insufficient capacity for vision-dense, high-resolution images, while wasting computation on vision-easy, low-resolution images; (2) pooling intermediate process images into a fixed token length during training may discard spatial details, making it harder for the model to learn semantically meaningful latent embeddings. 

As shown in Figure~\ref{Fig.dynamic}, SwimBird addresses these issues with a resolution-aware, dynamic latent token budget. Benefiting from the naive-resolution property of the Qwen ViT, we assign different maximum pixel budgets to the question image and the intermediate thinking images during training, which directly controls the maximum number of visual tokens produced by the vision encoder for each type of image. Concretely, we allow the vision encoder to output a variable number of visual tokens according to image resolution, bounded by an independent range $[N_{\min}, N_{\max}]$ (implemented via pixel/patch budget control). This avoids aggressive pooling that discards fine-grained evidence, while preventing excessively long visual sequences from dominating computation. Consequently, SwimBird can preserve detailed visual information when needed (e.g., tiny targets or dense diagrams) and remain efficient on simpler cases.

With this resolution-aware training setup, SwimBird further learns to allocate latent computation dynamically at inference time. In vision-only and interleaved modes, the number of latent tokens $K$ is not pre-defined: the model keeps generating latent embeddings until it decides to stop by emitting \texttt{</latent>}. This variable-length latent span naturally matches the amount of visual thinking to the perceived difficulty of the query.

\subsection{Switchable Reasoning SFT Dataset Construction}

To enable switchable reasoning modes, we curate a diverse SFT dataset covering three reasoning patterns:  (1) text-only CoT, (2) vision-only CoT where intermediate images are sufficient, and (3) interleaved vision-text CoT requiring both modalities. Our curation pipeline consists of three stages:

\textbf{Stage 1: Candidate collection and easy-instance filtering.}
We collect raw image-text interleaved CoT data from ThinkMorph~\cite{gu2025thinkmorph}, Zebra-CoT~\cite{li2025zebra}, and MathCanvas-Instruct~\cite{shi2025mathcanvas}. These datasets provide multimodal reasoning chains with intermediate visual thinking steps. where each sample contains intermediate thinking images. To focus on cases where intermediate visual reasoning is useful, we remove instances that are already solvable from the original input: Qwen3VL-8B is evaluated on the question and the original image, and correctly answered samples are filtered out.

\textbf{Stage 2: Reasoning-mode labeling via pass@8.} 
For each remaining sample, we compute two pass@8 scores with Qwen3VL-8B: $\text{pass}_{\text{base}}$ using only the question and problem image, and $\text{pass}_{\text{hint}}$ additionally providing the intermediate thinking images as visual hints. We judge each sampled answer using Qwen3-235B-Instruct given the question, prediction, and ground truth. We keep samples with $\text{pass}_{\text{hint}} \ge \text{pass}_{\text{base}}$, indicating that intermediate thinking images provide non-negative gains. Among them, we label samples with $\text{pass}_{\text{hint}} \ge 0.75$ as vision-only, since the model can solve the problem with high probability using the intermediate thinking images without an explicit textual CoT. The remaining kept samples, where $\text{pass}_{\text{hint}} \ge \text{pass}_{\text{base}}$ but $\text{pass}_{\text{hint}} < 0.75$, are labeled as interleaved vision--text, since the images help but are insufficient for consistently correct solutions and textual reasoning is still needed. This procedure yields 42K high-quality SFT samples covering the vision-only and interleaved modes.

\textbf{Stage 3: Add text-only CoT data.}
To complete the three-mode training set, we sample 50K text-only CoT instances from OpenMMReasoner~\cite{zhang2025openmmreasoner}, which provides pass@8-filtered textual CoT traces. Combining them with the 42K samples from Stage~2 yields \textbf{SwimBird-SFT-92K}, covering text-only, vision-only, and interleaved vision--text patterns. Detailed statistics are reported in Table~\ref{tab:dataset}.

\begin{table*}[t]
	\centering
	\resizebox{\linewidth}{!}{
		\begin{tabular}{l | cccc|c}
			\toprule
			\textbf{Model} & \textbf{V* Bench} &  \textbf{HR-Bench 4K} &  \textbf{HR-Bench 8K} & \textbf{MME RealWorld} & \textbf{Avg.} 
			\\
			\midrule
			\multicolumn{6}{c}{\textit{\textbf{Textual Reasoning Models}}} \\
			\midrule
			GPT-4o~\citep{hurst2024gpt} &66.0 &59.0 &55.5 &62.8 &60.9 \\
			GPT-5-mini &63.9 &66.3&60.9&- &- \\
			Qwen2.5-VL-32B-Instruct & 80.6 & 69.3 & 63.6 & 59.1 & 68.2 \\
			Qwen2.5-VL-7B-Instruct& 75.3 & 65.5 & 62.1 & 56.8 & 64.9 \\
			Qwen3-VL-8B-Instruct * & 83.8 & 76.5 & 71.3 & 61.9 & 73.4 \\
			Qwen3-VL-8B-Thinking & 77.5 & 72.4 & 68.1 &- &- \\
			InternVL3-8B~\citep{zhu2025internvl3} &81.2 &70.0 &69.3 &-&- \\
			LLaVA-OneVison~\citep{li2024llava} & 75.4 & 63.0 & 59.8 & 57.4 & 63.9 \\
			Vision-R1~\cite{huang2025vision} &80.1&64.8&57.0&- &-\\
			\midrule
			\multicolumn{6}{c}{\textit{\textbf{Latent Visual Reasoning Models}}} \\
			\midrule
			Monet~\citep{wang2025monet} & 83.3 & 71.0 & 68.0 & - & - \\
			LVR~\cite{li2025latent} &81.7 &69.6&66.1&-&-\\
			SkiLa~\cite{tong2025sketch} &84.3&72.0&66.5&-&- \\
			\midrule
			\multicolumn{6}{c}{\textit{\textbf{Multimodal Agentic Models}}} \\
			\midrule
			SEAL~\citep{wu2024v} & 74.8 & - & - & - & - \\
			Pixel Reasoner~\citep{wang2025pixel}  & 84.3 & 72.6 & 66.1 & 64.4 & 71.9 \\
			DeepEyes~\citep{zheng2025deepeyes}  & 83.3 & 73.2 & 69.5 & 64.1 & 72.5 \\
			Thyme~\cite{zhang2025thyme} & 82.2 & 77.0 & 72.0 & 64.8 & 74.0 \\
			DeepEyesV2~\cite{hong2025deepeyesv2}  & 81.8 & 77.9 & 73.8 & 64.9 & 74.6 \\
			\midrule
			\rowcolor{blue!8}
			\textbf{SwimBird} &85.5 & 79.0 & 74.9 &65.3 &76.2 \\
			\bottomrule
		\end{tabular}
	} 
	\caption{Performance on fine-grained visual understanding benchmarks. Here, * denotes the results are reproduced by ourselves.}
	\label{tab:perception} 
\end{table*}

\section{Experiments}

\textbf{Training Details}
We adopt Qwen3-VL 8B~\cite{bai2025qwen2} as the base model and conduct supervised fine-tuning on our curated SwimBird-SFT-92K. Training is performed on A100-80G GPUs with a global batch size of 128. The vision encoder and multimodal projector are kept frozen, and only the LLM parameters are updated. A cosine learning rate scheduler is applied with an initial learning rate of 1e-5.

\textbf{Baselines and Benchmarks}
To comprehensively assess the effectiveness of SwimBird, we compare it against three categories of baselines: (1) textual reasoning models, including advanced closed-source systems (e.g., GPT-4o and GPT-5-mini) and state-of-the-art open-source models (e.g., Qwen2.5/3-VL, LLaVA-OneVision); (2) latent visual reasoning models (e.g., Monet, LVR, SkiLa); and (3) multimodal agentic models that rely on explicit tool/workflow designs (e.g., Pixel Reasoner, DeepEyes, Thyme). We evaluate on two groups of benchmarks: (i) fine-grained/high-resolution visual understanding (V* Bench~\cite{wu2024v}, HR-Bench 4K/8K~\cite{wang2025divide}, MME-RealWorld~\cite{zhang2024mme}; Table~\ref{tab:perception}), and (ii) general VQA and multimodal reasoning (MMStar~\cite{chen2024we}, RealWorldQA~\cite{dsouza2025comparative}, WeMath~\cite{qiao2025we}, DynaMath~\cite{zou2024dynamath}, MathVerse\_MINI~\cite{zhang2024mathverse}; Table~\ref{tab:reasoning}). Results marked with * are reproduced by ourselves.

\subsection{Main Results}

\textbf{Fine-grained Visual Understanding}
Table~\ref{tab:perception} demonstrates that SwimBird achieves state-of-the-art performance on fine-grained and high-resolution perception. SwimBird obtains 85.5 on V* Bench, 79.0 on HR-Bench 4K, and 74.9 on HR-Bench 8K, outperforming strong textual reasoning baselines such as Qwen3-VL-8B-Instruct (83.8/76.5/71.3). Notably, Qwen3-VL-Thinking performs worse than Qwen3-VL-Instruct on visual perception, further supporting our claim that a mismatched reasoning mode can harm performance. Furthermore, SwimBird also outperforms current state-of-the-art multimodal agentic models such as Thyme (82.2/77.0/72.0) and DeepEyesV2 (81.8/77.9/73.8), which enhance perception via explicit cropping tools, highlighting that SwimBird can achieve stronger fine-grained perception without relying on complex tool pipelines.
We attribute these gains to SwimBird’s query-adaptive reasoning mode switching and adaptive latent-token allocation. Fine-grained visual tasks often require precise spatial evidence that is difficult to faithfully compress into text; meanwhile, forcing latent visual thoughts on text-centric steps can be redundant. By switching to vision-only reasoning when dense perception is needed (and allocating more latent computation for high-resolution inputs), SwimBird better preserves visual details and reduces modality mismatch, leading to consistently higher accuracy.

\textbf{General VQA and Multimodal Reasoning}
Beyond perception, SwimBird also shows strong improvements on general VQA and reasoning-heavy benchmarks. As shown in Table~\ref{tab:reasoning}, SwimBird reaches 71.2 on MMStar and 73.1 on RealWorldQA, exceeding Qwen3-VL-8B-Instruct* (64.7/71.8) and even outperforming Qwen2.5-VL-32B-Instruct on MMStar. More importantly, SwimBird delivers clear gains on multimodal reasoning: 49.5 on WeMath, 67.2 on DynaMath, and 65.8 on MathVerse\_MINI, outperforming strong open-source methods and agentic models.
These results suggest that SwimBird’s latent visual thoughts do not come at the cost of symbolic reasoning. Instead, SwimBird stays in text-only reasoning when the task is primarily linguistic or mathematical, and invokes vision-only or interleaved latent thinking only when additional visual evidence is beneficial. Learned from the multi-pattern supervision in SwimBird-SFT-92K, this query-adaptive selection avoids redundant visual thoughts that could interfere with textual logic, while still leveraging latent visual computation for vision-dependent subproblems.

\begin{table*}[t]
	\centering
	\setlength{\tabcolsep}{5pt}
	\resizebox{\linewidth}{!}{
		\begin{tabular}{l | cc|ccc}
			\toprule
			\multirow{2}{*}{\textbf{Models}}  &\multicolumn{2}{c}{\textbf{General VQA}}&\multicolumn{3}{|c}{\textbf{Multimodal Reasoning}}\\
			& MMStar &RealWorldQA &WeMath &  DynaMath &  MathVerse\_MINI
			\\
			\midrule
			Qwen2.5-VL-32B-Instruct &70.3 &- &- &-&48.5 \\
			Qwen2.5-VL-7B-Instruct&60.3 &67.4 &34.6&53.3 &45.6 \\
			Qwen3-VL-8B-Instruct * &64.7 &71.8 &38.8 &65.3 &61.3 \\
			LLaVA-OneVision~\citep{li2024llava} &61.9&69.9&20.9&-&19.3\\
			DeepEyes~\citep{zheng2025deepeyes}  &- &- &38.9 &55.0&47.3  \\
			DeepEyesV2~\cite{hong2025deepeyesv2}  &- &- &38.1 &57.2 &52.7 \\
			SkiLa~\citep{tong2025sketch} &64.8 &69.3 &- &- &- \\
			\midrule
			\rowcolor{blue!8}
			\textbf{SwimBird} &71.2 &73.1 &49.5 &67.2 &65.8 \\
			\bottomrule
		\end{tabular}
	} 
	\caption{Performance on general vqa and multimodal reasoning tasks. Here, * denotes the results are reproduced by ourselves.}
	\label{tab:reasoning}
\end{table*}

\subsection{Ablation Studies}

\textbf{Impact of the Maximum Latent Token Budget.}
We study how the maximum latent token budget $N_{\max}$ influences performance under our dynamic range setting $[N_{\min},N_{\max}]$. We fix $N_{\min}=2$ to ensure small images can be encoded without losing effective resolution, and vary $N_{\max}\in\{16,32,64,128\}$. As shown in Table~\ref{tab:tokens}, increasing $N_{\max}$ from 16 to 32 yields clear gains on vision-dense benchmarks (HRBench4K: 76.4 vs.\ 79.0; HRBench8K: 71.4 vs.\ 74.9), indicating that a moderate upper bound provides sufficient capacity for high-resolution perception. However, further expanding $N_{\max}$ to 64 or 128 does not help and even degrades performance (e.g., HRBench8K: 74.9 vs.\ 73.4 vs.\ 71.8), while RealWorldQA slightly drops (73.1 vs.\ 72.7). This suggests that an overly large latent budget may introduce redundant visual computation and interfere with overall reasoning. Overall, $N_{\max}=32$ offers the best trade-off and is used as the default setting.

\begin{table}[!t]
	\begin{minipage}{0.49\linewidth}
	\centering
	\setlength{\tabcolsep}{5pt}
	\resizebox{\linewidth}{!}{
		\begin{tabular}{c | ccc}
			\toprule
			\textbf{Latent Tokens} & \textbf{HRBench4K} &\textbf{HRBench8K} &\textbf{RealWorldQA} \\
			\midrule
			16 &76.4 &71.4 &73.1 \\
			32&79.0 &74.9 &73.1 \\
			64 &77.8 &73.4 &72.7  \\
			128&76.0&71.8&72.7\\
			\bottomrule
		\end{tabular}
	}
	\caption{Impact of maximum latent tokens budget.}
	\label{tab:tokens} 
	\end{minipage}
	\hfill
	\begin{minipage}{0.49\linewidth}
	\centering
	\setlength{\tabcolsep}{5pt}
	\resizebox{\linewidth}{!}{
		\begin{tabular}{c | ccc}
			\toprule
			\textbf{MSE Weight} & \textbf{HRBench4K} &\textbf{HRBench8K} &\textbf{RealWorldQA} \\
			\midrule
			0.1 &79.0 &71.8 &72.8 \\
			0.2&79.0 &74.9 &73.1 \\
			0.5 &77.8 &75.9 &72.0  \\
			1.0 &79.4&73.8&71.9\\
			\bottomrule
		\end{tabular}
	} 
	\caption{Impact of MSE loss weight coefficients.} 
	\label{tab:weight} 
	\end{minipage}
\end{table}

\textbf{Impact of the MSE Loss Weight Coefficient.}
We ablate the weight of the visual-thought reconstruction loss by varying $\lambda_{\text{vis}}$ while keeping other settings fixed. As shown in Table~\ref{tab:weight}, a moderate MSE weight yields the most balanced performance. Specifically, setting $\lambda_{\text{vis}}=0.2$ achieves strong results across all benchmarks. When $\lambda_{\text{vis}}$ is too small (0.1), the supervision on latent visual thoughts becomes weak, leading to a notable drop on the most vision-dense benchmark (HRBench8K: 71.8). In contrast, increasing $\lambda_{\text{vis}}$ to 0.5 improves HRBench8K (75.9) but degrades RealWorldQA (72.0), suggesting that overly emphasizing MSE training may bias the model toward visual reconstruction at the expense of general multimodal reasoning. With $\lambda_{\text{vis}}=1.0$, HRBench4K slightly increases (79.4) but performance drops on HRBench8K and RealWorldQA, indicating instability under overly strong visual-loss weighting. Overall, we use $\lambda_{\text{vis}}=0.2$ as the default, which best balances visual reasoning and text-centric reasoning.

\begin{table}[!t]

\end{table}

\begin{figure}[!t]
	\centering
	\includegraphics[width=\linewidth]{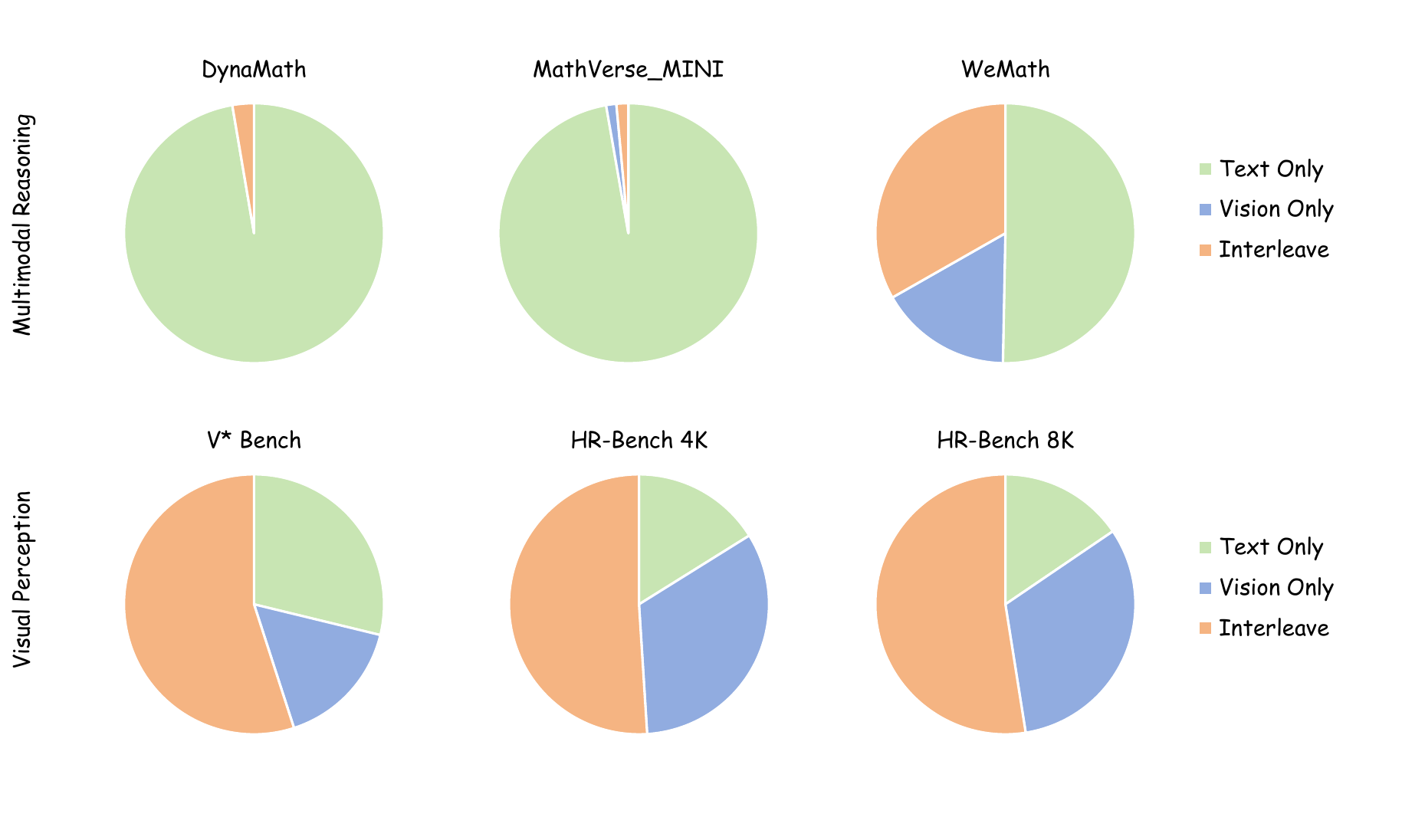} 
	\caption{\textbf{Distribution of reasoning mode across different benchmarks for SwimBird.}}
	\label{Fig.distribute} 
\end{figure}

\begin{figure*}[!t]
	\centering
	\includegraphics[width=\linewidth]{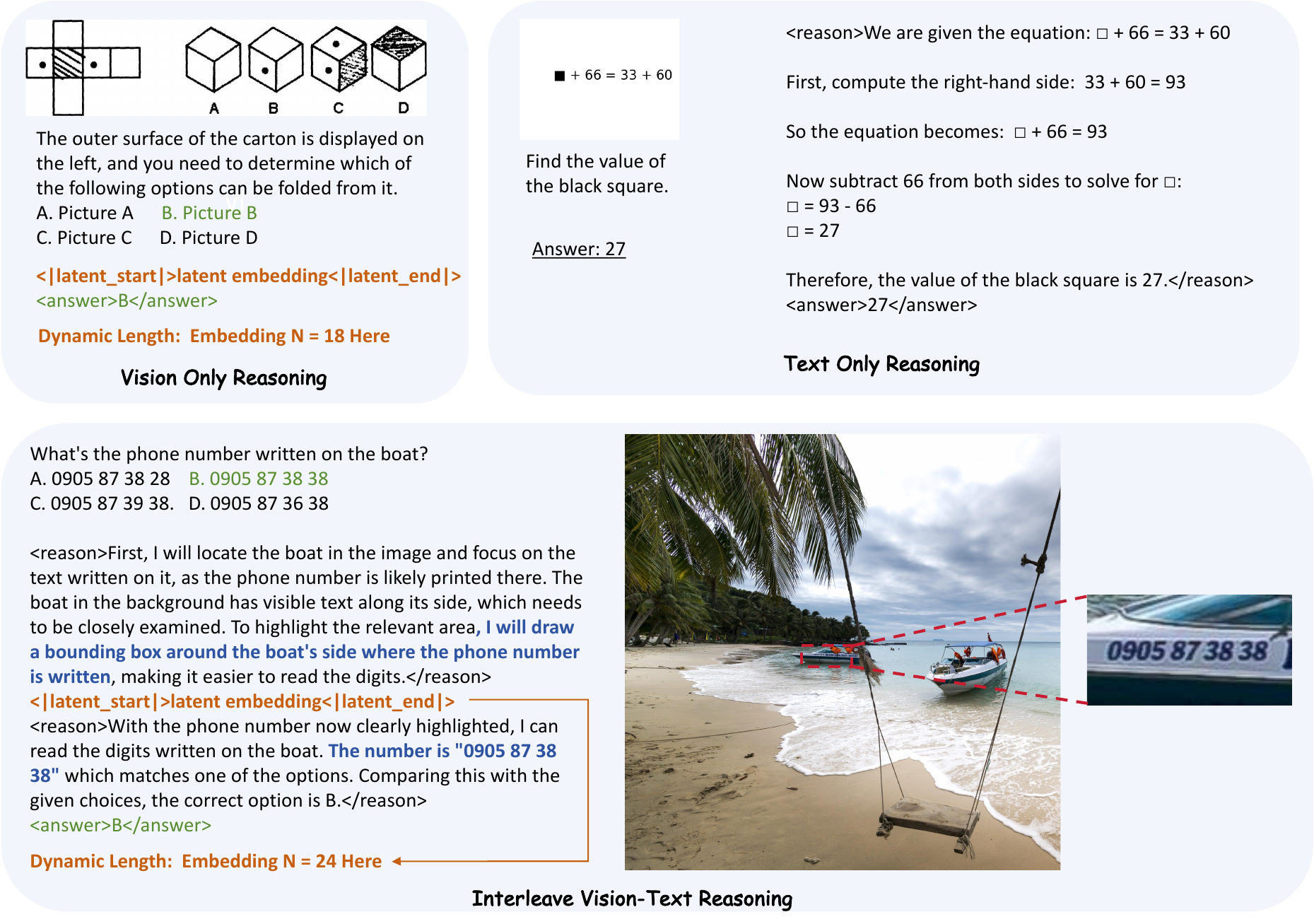} 
	\caption{\textbf{Analysis of  Different Reasoning-Mode Case.}}
	\label{Fig.case}
\end{figure*} 

\subsection{Analysis of Switchable Reasoning Mode}
\textbf{Analysis of Reasoning-Mode Distribution}
We analyze the distribution of SwimBird’s reasoning modes across benchmarks (Fig.~\ref{Fig.distribute}) to verify its query-adaptive behavior. Overall, the selected mode matches each benchmark’s dominant difficulty.
On \textbf{text-logic-dominant} multimodal reasoning datasets (DynaMath and MathVerse\_MINI), SwimBird almost always uses text-only reasoning, with vision-only and interleaved traces rarely triggered, suggesting it avoids redundant latent visual thoughts when symbolic manipulation and linguistic deduction are sufficient.
On \textbf{vision-dense perception} benchmarks (V* Bench and HR-Bench 4K/8K), SwimBird frequently activates vision-only and especially interleaved vision--text reasoning, reflecting the need to alternate between visual grounding (e.g., tiny targets in high-resolution images) and explicit textual deduction. The proportion of vision-only reasoning increases from HR-Bench 4K to 8K, consistent with higher perceptual load at higher resolutions.
WeMath exhibits a more \textbf{balanced} mixture of all three modes, where some problems are text-centric while others require substantial visual grounding. These results confirm that SwimBird does not follow a fixed template, but instead selects reasoning modes in an instance-dependent manner to mitigate modality mismatch.

\textbf{Analysis of Different Reasoning-Mode Cases}
Fig.~\ref{Fig.case} provides qualitative examples of SwimBird’s mode selection.
For vision-only reasoning (top-left), the cube-net folding problem mainly requires spatial perception and mental rotation; SwimBird directly enters a latent visual-thought span and outputs the answer without unnecessary textual CoT, while allocating an appropriate latent length (e.g., $N{=}18$).
For text-only reasoning (top-right), the arithmetic equation is purely symbolic; SwimBird solves it with textual deduction, avoiding redundant visual thoughts that could interfere with logical steps.
For interleaved vision--text reasoning (bottom), reading a phone number from a small region in a natural image requires both precise visual localization and explicit option comparison; SwimBird first uses latent visual thoughts to focus on the relevant region, then switches back to text for verification and decision making, again with a dynamically allocated latent length (e.g., $N{=}24$).
Together, these cases show that SwimBird mitigates modality mismatch by choosing when to think in vision versus language and by adaptively allocating visual-thought computation to match perceptual difficulty.

\section{Prompt}
\begin{figure*}[!h]
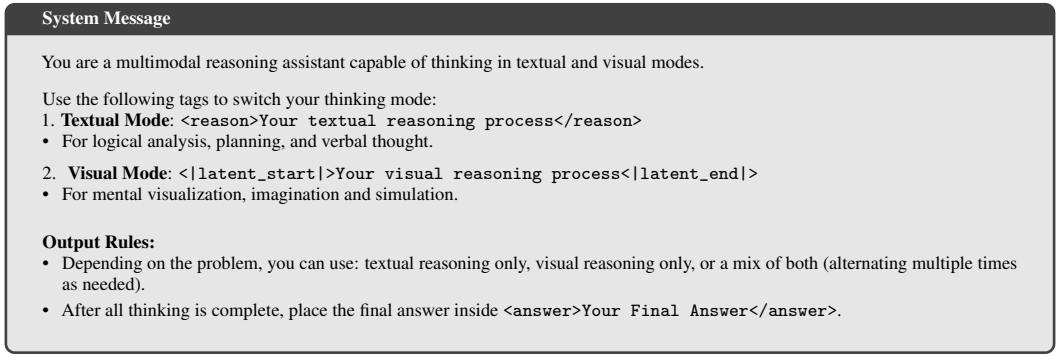

	\centering
	
	\begin{subfigure}{\textwidth}
		\begin{tcolorbox}[
			boxrule=1pt,
			boxsep=2pt,
			colback=gray!20,
			fontupper=\scriptsize,
			fonttitle=\scriptsize\bfseries,
			title=System Message
			]
			
			\setlist[itemize]{leftmargin=*, nosep, after=\vspace{4pt}} 
			
			You are a multimodal reasoning assistant capable of thinking in textual and visual modes.
			
			\medskip
			Use the following tags to switch your thinking mode:
			
			1.  \textbf{Textual Mode}:  \texttt{<reason>Your textual reasoning process</reason>}
			\begin{itemize}[leftmargin=*]
				\setlength{\itemsep}{1pt}
				\setlength{\parskip}{1pt}
				\item For logical analysis, planning, and verbal thought.
			\end{itemize}
			
			2. \textbf{ Visual Mode}: \texttt{<|latent\_start|>Your visual reasoning process<|latent\_end|>}
			\begin{itemize}[leftmargin=*]
				\setlength{\itemsep}{1pt}
				\setlength{\parskip}{1pt}
				\item For mental visualization, imagination and simulation.
			\end{itemize}
			\medskip
			
			\textbf{Output Rules:}
			\begin{itemize}[leftmargin=*]
				\setlength{\itemsep}{1pt}
				\setlength{\parskip}{1pt}
				\item Depending on the problem, you can use: textual reasoning only, visual reasoning only, or a mix of both (alternating multiple times as needed).
				\item After all thinking is complete, place the final answer inside \texttt{<answer>Your Final Answer</answer>}.
			\end{itemize}
		\end{tcolorbox}
	\end{subfigure}
	
	\caption{\textbf{System prompt used for SwimBird.}}
	\label{fig:rag_prompt}
\end{figure*}

To guide SwimBird's query-adaptive reasoning mode selection, we design a system prompt that explicitly instructs the model on how to switch between textual and visual thinking modes. As shown in Figure~\ref{fig:rag_prompt}, the prompt defines three reasoning patterns: text-only, vision-only, and interleaved, using structured tags (\texttt{<reason>} for textual thoughts and \texttt{<latent>} for visual thoughts), and allows the model to dynamically choose the most appropriate mode or combination based on the input query.

\section{Conclusion}
We present \textbf{SwimBird}, a reasoning-switchable MLLM that addresses the fixed reasoning pattern in prior multimodal CoT frameworks. SwimBird adopts a hybrid autoregressive paradigm and can adaptively switch among text-only, vision-only, and interleaved vision--text reasoning, while dynamically allocating the latent visual token budget. We also construct \textbf{SwimBird-SFT-92K} with a systematic curation and mode-labeling strategy to enable effective multi-mode training. Extensive experiments show that SwimBird achieves SoTA performance on both text-centric reasoning and challenging vision-dense tasks.

{
\small
\bibliographystyle{plain}
\bibliography{ref}

@article{kojima2022large,
	title={Large language models are zero-shot reasoners},
	author={Kojima, Takeshi and Gu, Shixiang Shane and Reid, Machel and Matsuo, Yutaka and Iwasawa, Yusuke},
	journal={Advances in neural information processing systems},
	volume={35},
	pages={22199--22213},
	year={2022}
}

@article{gu2025thinkmorph,
  title={ThinkMorph: Emergent Properties in Multimodal Interleaved Chain-of-Thought Reasoning},
  author={Gu, Jiawei and Hao, Yunzhuo and Wang, Huichen Will and Li, Linjie and Shieh, Michael Qizhe and Choi, Yejin and Krishna, Ranjay and Cheng, Yu},
  journal={arXiv preprint arXiv:2510.27492},
  year={2025}
}

@article{li2025zebra,
  title={Zebra-cot: A dataset for interleaved vision language reasoning},
  author={Li, Ang and Wang, Charles and Fu, Deqing and Yue, Kaiyu and Cai, Zikui and Zhu, Wang Bill and Liu, Ollie and Guo, Peng and Neiswanger, Willie and Huang, Furong and others},
  journal={arXiv preprint arXiv:2507.16746},
  year={2025}
}

@article{shi2025mathcanvas,
  title={Mathcanvas: Intrinsic visual chain-of-thought for multimodal mathematical reasoning},
  author={Shi, Weikang and Yu, Aldrich and Fang, Rongyao and Ren, Houxing and Wang, Ke and Zhou, Aojun and Tian, Changyao and Fu, Xinyu and Hu, Yuxuan and Lu, Zimu and others},
  journal={arXiv preprint arXiv:2510.14958},
  year={2025}
}

@article{yan2025crosslmm,
  title={Crosslmm: Decoupling long video sequences from lmms via dual cross-attention mechanisms},
  author={Yan, Shilin and Han, Jiaming and Tsai, Joey and Xue, Hongwei and Fang, Rongyao and Hong, Lingyi and Guo, Ziyu and Zhang, Ray},
  journal={arXiv preprint arXiv:2505.17020},
  year={2025}
}

@article{wei2022chain,
	title={Chain-of-thought prompting elicits reasoning in large language models},
	author={Wei, Jason and Wang, Xuezhi and Schuurmans, Dale and Bosma, Maarten and Xia, Fei and Chi, Ed and Le, Quoc V and Zhou, Denny and others},
	journal={Advances in neural information processing systems},
	volume={35},
	pages={24824--24837},
	year={2022}
}

@article{wang2024qwen2,
	title={Qwen2-vl: Enhancing vision-language model's perception of the world at any resolution},
	author={Wang, Peng and Bai, Shuai and Tan, Sinan and Wang, Shijie and Fan, Zhihao and Bai, Jinze and Chen, Keqin and Liu, Xuejing and Wang, Jialin and Ge, Wenbin and others},
	journal={arXiv preprint arXiv:2409.12191},
	year={2024}
}

@article{bai2025qwen2,
	title={Qwen2. 5-vl technical report},
	author={Bai, Shuai and Chen, Keqin and Liu, Xuejing and Wang, Jialin and Ge, Wenbin and Song, Sibo and Dang, Kai and Wang, Peng and Wang, Shijie and Tang, Jun and others},
	journal={arXiv preprint arXiv:2502.13923},
	year={2025}
}

@article{wang2025internvl3,
	title={Internvl3. 5: Advancing open-source multimodal models in versatility, reasoning, and efficiency},
	author={Wang, Weiyun and Gao, Zhangwei and Gu, Lixin and Pu, Hengjun and Cui, Long and Wei, Xingguang and Liu, Zhaoyang and Jing, Linglin and Ye, Shenglong and Shao, Jie and others},
	journal={arXiv preprint arXiv:2508.18265},
	year={2025}
}

@article{zhu2025internvl3,
	title={Internvl3: Exploring advanced training and test-time recipes for open-source multimodal models},
	author={Zhu, Jinguo and Wang, Weiyun and Chen, Zhe and Liu, Zhaoyang and Ye, Shenglong and Gu, Lixin and Tian, Hao and Duan, Yuchen and Su, Weijie and Shao, Jie and others},
	journal={arXiv preprint arXiv:2504.10479},
	year={2025}
}

@article{liu2023visual,
	title={Visual instruction tuning},
	author={Liu, Haotian and Li, Chunyuan and Wu, Qingyang and Lee, Yong Jae},
	journal={Advances in neural information processing systems},
	volume={36},
	pages={34892--34916},
	year={2023}
}

@article{zhang2025openmmreasoner,
	title={Openmmreasoner: Pushing the frontiers for multimodal reasoning with an open and general recipe},
	author={Zhang, Kaichen and Wu, Keming and Yang, Zuhao and Li, Bo and Hu, Kairui and Wang, Bin and Liu, Ziwei and Li, Xingxuan and Bing, Lidong},
	journal={arXiv preprint arXiv:2511.16334},
	year={2025}
}

@misc{liu2024llavanext,
	title={LLaVA-NeXT: Improved reasoning, OCR, and world knowledge},
	url={https://llava-vl.github.io/blog/2024-01-30-llava-next/},
	author={Liu, Haotian and Li, Chunyuan and Li, Yuheng and Li, Bo and Zhang, Yuanhan and Shen, Sheng and Lee, Yong Jae},
	month={January},
	year={2024}
}

@inproceedings{xu2025llava,
	title={Llava-cot: Let vision language models reason step-by-step},
	author={Xu, Guowei and Jin, Peng and Wu, Ziang and Li, Hao and Song, Yibing and Sun, Lichao and Yuan, Li},
	booktitle={Proceedings of the IEEE/CVF International Conference on Computer Vision},
	pages={2087--2098},
	year={2025}
}

@inproceedings{qiao2025we,
	title={We-math: Does your large multimodal model achieve human-like mathematical reasoning?},
	author={Qiao, Runqi and Tan, Qiuna and Dong, Guanting and MinhuiWu, MinhuiWu and Sun, Chong and Song, Xiaoshuai and Wang, Jiapeng and Gongque, Zhuoma and Lei, Shanglin and Zhang, Yifan and others},
	booktitle={Proceedings of the 63rd Annual Meeting of the Association for Computational Linguistics (Volume 1: Long Papers)},
	pages={20023--20070},
	year={2025}
}

@inproceedings{yue2024mmmu,
	title={Mmmu: A massive multi-discipline multimodal understanding and reasoning benchmark for expert agi},
	author={Yue, Xiang and Ni, Yuansheng and Zhang, Kai and Zheng, Tianyu and Liu, Ruoqi and Zhang, Ge and Stevens, Samuel and Jiang, Dongfu and Ren, Weiming and Sun, Yuxuan and others},
	booktitle={Proceedings of the IEEE/CVF Conference on Computer Vision and Pattern Recognition},
	pages={9556--9567},
	year={2024}
}

@article{wang2024charxiv,
	title={Charxiv: Charting gaps in realistic chart understanding in multimodal llms},
	author={Wang, Zirui and Xia, Mengzhou and He, Luxi and Chen, Howard and Liu, Yitao and Zhu, Richard and Liang, Kaiqu and Wu, Xindi and Liu, Haotian and Malladi, Sadhika and others},
	journal={Advances in Neural Information Processing Systems},
	volume={37},
	pages={113569--113697},
	year={2024}
}

@inproceedings{li2023blip,
	title={Blip-2: Bootstrapping language-image pre-training with frozen image encoders and large language models},
	author={Li, Junnan and Li, Dongxu and Savarese, Silvio and Hoi, Steven},
	booktitle={International conference on machine learning},
	pages={19730--19742},
	year={2023},
	organization={PMLR}
}

@article{zou2024dynamath,
  title={Dynamath: A dynamic visual benchmark for evaluating mathematical reasoning robustness of vision language models},
  author={Zou, Chengke and Guo, Xingang and Yang, Rui and Zhang, Junyu and Hu, Bin and Zhang, Huan},
  journal={arXiv preprint arXiv:2411.00836},
  year={2024}
}

@article{zhang2023multimodal,
	title={Multimodal chain-of-thought reasoning in language models},
	author={Zhang, Zhuosheng and Zhang, Aston and Li, Mu and Zhao, Hai and Karypis, George and Smola, Alex},
	journal={arXiv preprint arXiv:2302.00923},
	year={2023}
}

@article{huang2025vision,
	title={Vision-r1: Incentivizing reasoning capability in multimodal large language models},
	author={Huang, Wenxuan and Jia, Bohan and Zhai, Zijie and Cao, Shaosheng and Ye, Zheyu and Zhao, Fei and Xu, Zhe and Hu, Yao and Lin, Shaohui},
	journal={arXiv preprint arXiv:2503.06749},
	year={2025}
}

@article{liu2025visual,
	title={Visual-rft: Visual reinforcement fine-tuning},
	author={Liu, Ziyu and Sun, Zeyi and Zang, Yuhang and Dong, Xiaoyi and Cao, Yuhang and Duan, Haodong and Lin, Dahua and Wang, Jiaqi},
	journal={arXiv preprint arXiv:2503.01785},
	year={2025}
}

@inproceedings{zhang2024mathverse,
  title={Mathverse: Does your multi-modal llm truly see the diagrams in visual math problems?},
  author={Zhang, Renrui and Jiang, Dongzhi and Zhang, Yichi and Lin, Haokun and Guo, Ziyu and Qiu, Pengshuo and Zhou, Aojun and Lu, Pan and Chang, Kai-Wei and Qiao, Yu and others},
  booktitle={European Conference on Computer Vision},
  pages={169--186},
  year={2024},
  organization={Springer}
}

@article{yu2025perception,
	title={Perception-r1: Pioneering perception policy with reinforcement learning},
	author={Yu, En and Lin, Kangheng and Zhao, Liang and Yin, Jisheng and Wei, Yana and Peng, Yuang and Wei, Haoran and Sun, Jianjian and Han, Chunrui and Ge, Zheng and others},
	journal={arXiv preprint arXiv:2504.07954},
	year={2025}
}

@article{dsouza2025comparative,
  title={Comparative Analysis of Leading Generative AI Conversational Systems: ChatGPT, Grok AI, Gemini, and Meta AI},
  author={Dsouza, Nigel},
  journal={Authorea Preprints},
  year={2025},
  publisher={Authorea}
}

@inproceedings{fu2024blink,
	title={Blink: Multimodal large language models can see but not perceive},
	author={Fu, Xingyu and Hu, Yushi and Li, Bangzheng and Feng, Yu and Wang, Haoyu and Lin, Xudong and Roth, Dan and Smith, Noah A and Ma, Wei-Chiu and Krishna, Ranjay},
	booktitle={European Conference on Computer Vision},
	pages={148--166},
	year={2024},
	organization={Springer}
}

@article{yang2025machine,
	title={Machine Mental Imagery: Empower Multimodal Reasoning with Latent Visual Tokens},
	author={Yang, Zeyuan and Yu, Xueyang and Chen, Delin and Shen, Maohao and Gan, Chuang},
	journal={arXiv preprint arXiv:2506.17218},
	year={2025}
}

@article{li2025latent,
	title={Latent visual reasoning},
	author={Li, Bangzheng and Sun, Ximeng and Liu, Jiang and Wang, Ze and Wu, Jialian and Yu, Xiaodong and Chen, Hao and Barsoum, Emad and Chen, Muhao and Liu, Zicheng},
	journal={arXiv preprint arXiv:2509.24251},
	year={2025}
}

@article{tong2025sketch,
	title={Sketch-in-latents: Eliciting unified reasoning in mllms},
	author={Tong, Jintao and Gu, Jiaqi and Lou, Yujing and Fan, Lubin and Zou, Yixiong and Wu, Yue and Ye, Jieping and Li, Ruixuan},
	journal={arXiv preprint arXiv:2512.16584},
	year={2025}
}

@article{zheng2025deepeyes,
	title={DeepEyes: Incentivizing" Thinking with Images" via Reinforcement Learning},
	author={Zheng, Ziwei and Yang, Michael and Hong, Jack and Zhao, Chenxiao and Xu, Guohai and Yang, Le and Shen, Chao and Yu, Xing},
	journal={arXiv preprint arXiv:2505.14362},
	year={2025}
}

@article{zhang2025thyme,
	title={Thyme: Think beyond images},
	author={Zhang, Yi-Fan and Lu, Xingyu and Yin, Shukang and Fu, Chaoyou and Chen, Wei and Hu, Xiao and Wen, Bin and Jiang, Kaiyu and Liu, Changyi and Zhang, Tianke and others},
	journal={arXiv preprint arXiv:2508.11630},
	year={2025}
}

@inproceedings{wu2024v,
	title={V?: Guided visual search as a core mechanism in multimodal llms},
	author={Wu, Penghao and Xie, Saining},
	booktitle={Proceedings of the IEEE/CVF Conference on Computer Vision and Pattern Recognition},
	pages={13084--13094},
	year={2024}
}

@article{hong2025deepeyesv2,
	title={DeepEyesV2: Toward Agentic Multimodal Model},
	author={Hong, Jack and Zhao, Chenxiao and Zhu, ChengLin and Lu, Weiheng and Xu, Guohai and Yu, Xing},
	journal={arXiv preprint arXiv:2511.05271},
	year={2025}
}

@article{wang2025pixel,
	title={Pixel reasoner: Incentivizing pixel-space reasoning with curiosity-driven reinforcement learning},
	author={Wang, Haozhe and Su, Alex and Ren, Weiming and Lin, Fangzhen and Chen, Wenhu},
	journal={arXiv preprint arXiv:2505.15966},
	year={2025}
}

@inproceedings{tong2025emosync,
  title={EmoSync: Multi-Stage Reasoning with Multimodal Large Language Models for Fine-Grained Emotion Recognition},
  author={Tong, Jintao and Li, Shiwei and Zhuang, Zijian and Hu, Jinghan and Zou, Yixiong},
  booktitle={Proceedings of the 3rd International Workshop on Multimodal and Responsible Affective Computing},
  pages={95--99},
  year={2025}
}

@article{li2024llava,
	title={Llava-onevision: Easy visual task transfer},
	author={Li, Bo and Zhang, Yuanhan and Guo, Dong and Zhang, Renrui and Li, Feng and Zhang, Hao and Zhang, Kaichen and Zhang, Peiyuan and Li, Yanwei and Liu, Ziwei and others},
	journal={arXiv preprint arXiv:2408.03326},
	year={2024}
}

@inproceedings{shen2025zoomeye,
  title={Zoomeye: Enhancing multimodal llms with human-like zooming capabilities through tree-based image exploration},
  author={Shen, Haozhan and Zhao, Kangjia and Zhao, Tiancheng and Xu, Ruochen and Zhang, Zilun and Zhu, Mingwei and Yin, Jianwei},
  booktitle={Proceedings of the 2025 Conference on Empirical Methods in Natural Language Processing},
  pages={6613--6629},
  year={2025}
}

@article{hurst2024gpt,
	title={Gpt-4o system card},
	author={Hurst, Aaron and Lerer, Adam and Goucher, Adam P and Perelman, Adam and Ramesh, Aditya and Clark, Aidan and Ostrow, AJ and Welihinda, Akila and Hayes, Alan and Radford, Alec and others},
	journal={arXiv preprint arXiv:2410.21276},
	year={2024}
}

@article{zhang2025latent,
	title={Latent sketchpad: Sketching visual thoughts to elicit multimodal reasoning in mllms},
	author={Zhang, Huanyu and Wu, Wenshan and Li, Chengzu and Shang, Ning and Xia, Yan and Huang, Yangyu and Zhang, Yifan and Dong, Li and Zhang, Zhang and Wang, Liang and others},
	journal={arXiv preprint arXiv:2510.24514},
	year={2025}
}

@article{shen2025vlm,
  title={Vlm-r1: A stable and generalizable r1-style large vision-language model},
  author={Shen, Haozhan and Liu, Peng and Li, Jingcheng and Fang, Chunxin and Ma, Yibo and Liao, Jiajia and Shen, Qiaoli and Zhang, Zilun and Zhao, Kangjia and Zhang, Qianqian and others},
  journal={arXiv preprint arXiv:2504.07615},
  year={2025}
}

@article{wang2025vl,
	title={Vl-rethinker: Incentivizing self-reflection of vision-language models with reinforcement learning},
	author={Wang, Haozhe and Qu, Chao and Huang, Zuming and Chu, Wei and Lin, Fangzhen and Chen, Wenhu},
	journal={arXiv preprint arXiv:2504.08837},
	year={2025}
}

@article{chen2024we,
  title={Are we on the right way for evaluating large vision-language models?},
  author={Chen, Lin and Li, Jinsong and Dong, Xiaoyi and Zhang, Pan and Zang, Yuhang and Chen, Zehui and Duan, Haodong and Wang, Jiaqi and Qiao, Yu and Lin, Dahua and others},
  journal={Advances in Neural Information Processing Systems},
  volume={37},
  pages={27056--27087},
  year={2024}
}

@inproceedings{wang2025divide,
	title={Divide, conquer and combine: A training-free framework for high-resolution image perception in multimodal large language models},
	author={Wang, Wenbin and Ding, Liang and Zeng, Minyan and Zhou, Xiabin and Shen, Li and Luo, Yong and Yu, Wei and Tao, Dacheng},
	booktitle={Proceedings of the AAAI Conference on Artificial Intelligence},
	volume={39},
	number={8},
	pages={7907--7915},
	year={2025}
}

@article{zhang2024mme,
	title={Mme-realworld: Could your multimodal llm challenge high-resolution real-world scenarios that are difficult for humans?},
	author={Zhang, Yi-Fan and Zhang, Huanyu and Tian, Haochen and Fu, Chaoyou and Zhang, Shuangqing and Wu, Junfei and Li, Feng and Wang, Kun and Wen, Qingsong and Zhang, Zhang and others},
	journal={arXiv preprint arXiv:2408.13257},
	year={2024}
}

@article{wang2025monet,
	title={Monet: Reasoning in latent visual space beyond images and language},
	author={Wang, Qixun and Shi, Yang and Wang, Yifei and Zhang, Yuanxing and Wan, Pengfei and Gai, Kun and Ying, Xianghua and Wang, Yisen},
	journal={arXiv preprint arXiv:2511.21395},
	year={2025}
}

@article{tong2025flowcut,
  title={FlowCut: Rethinking Redundancy via Information Flow for Efficient Vision-Language Models},
  author={Tong, Jintao and Jin, Wenwei and Qin, Pengda and Li, Anqi and Zou, Yixiong and Li, Yuhong and Li, Yuhua and Li, Ruixuan},
  journal={arXiv preprint arXiv:2505.19536},
  year={2025}
}

@article{qin2025chain,
	title={Chain-of-Visual-Thought: Teaching VLMs to See and Think Better with Continuous Visual Tokens},
	author={Qin, Yiming and Wei, Bomin and Ge, Jiaxin and Kallidromitis, Konstantinos and Fu, Stephanie and Darrell, Trevor and Wang, XuDong},
	journal={arXiv preprint arXiv:2511.19418},
	year={2025}
}
}
%
%

\clearpage


\end{document}